\documentclass{article}

\usepackage[nonatbib, final]{neurips_2019}

\usepackage{natbib}
\usepackage[
  linktoc=all,
  pdfcenterwindow=true,
  bookmarks=true,
  colorlinks,
  citecolor=CiteColour,
  linkcolor=LinkColour,
  urlcolor=CommentDarkBlue]{hyperref}

\usepackage[utf8]{inputenc} % allow utf-8 input
\usepackage[T1]{fontenc}    % use 8-bit T1 fonts
\usepackage{hyperref}       % hyperlinks
\usepackage{url}            % simple URL typesetting
\usepackage{booktabs}       % professional-quality tables
\usepackage{amsfonts}       % blackboard math symbols
\usepackage{nicefrac}       % compact symbols for 1/2, etc.
\usepackage{microtype}      % microtypography

\usepackage{color}
\usepackage{amsmath}
\usepackage{bbm}
\usepackage{graphicx, caption, subcaption}
\usepackage{tabularx}
\usepackage{multirow}
\usepackage{breqn}
\DeclareMathOperator*{\argmax}{arg\,max}

\newcommand{\y}{\mathbf{y}} 
\newcommand{\x}{\mathbf{x}}
\newcommand{\z}{\mathbf{z}}
\newcommand{\bpi}{{\boldsymbol\pi}}
\newcommand{\kl}[2]{\textrm{KL}{\left( #1 \dline #2 \right)}}

\newcommand{\given}{\, | \,}
\newcommand{\dline}{\, || \,}

\newcommand{\prev}{_{prev}}

\newcommand{\tprev}{\theta\prev}

\newcommand{\pprev}{p_{\tprev}}

\newcommand{\Dexp}{\mathcal{D}_{new}}
\newcommand{\Nexp}{N_{new}}

\newcommand{\cexp}{c_{new}}

\usepackage[usenames,dvipsnames]{xcolor}
\definecolor{shadecolor}{gray}{0.9}
\colorlet{CiteColour}{black!40!blue!100}
\colorlet{LinkColour}{black!40!blue!100}
\definecolor{CommentDarkBlue}{rgb}{0,0,0.4}

\title{Continual Unsupervised Representation Learning}

\author{%
  Dushyant Rao, Francesco Visin, Andrei A. Rusu, \\ \textbf{Yee Whye Teh, Razvan Pascanu, Raia Hadsell}\thanks{Correspondence to: \texttt{\{dushyantr, visin\}@google.com}} \\
  DeepMind\\
  London, UK \\
}

\begin{document}

\maketitle

\begin{abstract}

Continual learning aims to improve the ability of modern learning systems to deal with non-stationary distributions, typically by attempting to learn a series of tasks sequentially. Prior art in the field has largely considered supervised or reinforcement learning tasks, and often assumes full knowledge of task labels and boundaries.
In this work, we propose an approach (CURL) to tackle a more general problem that we will refer to as \textit{unsupervised continual learning}. The focus is on learning representations without any knowledge about task identity, and we explore scenarios when there are abrupt changes between tasks, smooth transitions from one task to another, or even when the data is shuffled.
The proposed approach performs task inference directly within the model, is able to dynamically expand to capture new concepts over its lifetime, and incorporates additional rehearsal-based techniques to deal with catastrophic forgetting.
We demonstrate the efficacy of CURL in an unsupervised learning setting with MNIST and Omniglot, where the lack of labels ensures no information is leaked about the task.
Further, we demonstrate strong performance compared to prior art in an i.i.d setting, or when adapting the technique to supervised tasks such as incremental class learning.

\end{abstract}

\section{Introduction}
Humans have the impressive ability to learn many different concepts and perform different tasks in a sequential lifelong setting.
For example, infants learn to interact with objects in their environment without clear specification of tasks (\textit{task-agnostic}), in a sequential fashion without forgetting (\textit{non-stationary}), from temporally correlated visual inputs (\textit{non-i.i.d}), and with minimal external supervision (\textit{unsupervised}). 
For a learning system such as a robot deployed in the real world, it is highly desirable to satisfy these desiderata as well.
In contrast, learning algorithms often require input samples to be shuffled in order to satisfy the i.i.d. assumption, and have been shown to perform poorly when trained on sequential data, with newer tasks or concepts overwriting older ones; a phenomenon known as catastrophic forgetting~\citep{mccloskey1989catastrophic, goodfellow2013empirical}.
As a result, there has been renewed research focus on the \textit{continual learning} problem in recent years~\citep[e.g.][]{kirkpatrick2017overcoming, nguyen2017variational, zenke2017continual, shin2017continual},
with several approaches addressing catastrophic forgetting as well as backwards or forwards transfer---using the current task to improve performance on past or future tasks.
However, most of these techniques have focused on a sequence of tasks in which both the identity of the task (\emph{task label}) and boundaries between tasks are provided; moreover, they often focus on the supervised learning setting, where \emph{class labels} for each data point are given.
Thus, many of these methods fail to capture some of the aforementioned properties of real-world continual learning, with unknown task labels or poorly defined task boundaries, or when abundant class-labelled data is not available.
In this paper, we propose to address the more general \textit{unsupervised continual learning} setting (also suggested separately by~\citet{smith2019unsupervised}), in which task labels and boundaries are not provided to the learner, and hence the focus is on unsupervised task learning.
The tasks could correspond to either unsupervised representation learning, or learning skills without extrinsic reward if applied to the reinforcement learning domain.
In this sense, the problem setting is ``unsupervised'' in two ways: in terms of the absence of task labels (or indeed well-defined tasks themselves), and in terms of the absence of external supervision such as class labels, regression targets, or external rewards.
The two aspects may seem independent, but considering the unsupervised learning problem encourages solutions that aim to capture all fundamental properties of the data, which in turn might encourage, or reinforce, particular ways of addressing the task boundary problem. Hence the two aspects are connected through the type of solutions they necessitate, and it is beneficial to consider them jointly.
We argue that this is an important and challenging open problem, as it enables continual learning in environments without clearly defined tasks and goals, and with minimal external supervision.
Relaxing these constraints is crucial to performing lifelong learning in the real world.

Our approach, named \textit{Continual Unsupervised Representation Learning} (CURL), learns a task-specific representation on top of a larger set of shared parameters, and deals with task ambiguity by performing task inference within the model.
We endow the model with the ability to dynamically expand its capacity to capture new tasks, and suggest methods to minimise catastrophic forgetting.
The model is experimentally evaluated in a variety of unsupervised settings: when tasks or classes are presented sequentially, when training data are shuffled, and with ambiguous task boundaries when transitions are continuous rather than discrete.
We also demonstrate that despite focusing on unsupervised learning, the method can be trivially adapted to supervised learning while removing the reliance on task knowledge and class labels.
The experiments demonstrate competitive performance with respect to previous work, with the additional ability to learn without supervision in a continual learning setting, and indicate the efficacy of the different components of the proposed method.
\section{Model}

We begin by defining the CURL model and training loss, then introduce methods to perform dynamic expansion, and propose a generative replay mechanism to combat forgetting.

\begin{figure}[t]
\begin{minipage}[b]{0.3\textwidth}
  \centering
    \includegraphics[height=0.12\textheight,clip,trim=0 0 -5cm 0]{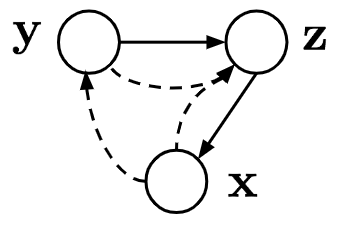}
    \caption{Graphical model for CURL. The categorical task variable $\y$ is used to instantiate a latent mixture-of-Gaussians $\z$, which is then decoded to $\x$.}
    \label{fig:pgm}
\end{minipage}
\hspace{0.5cm}
\begin{minipage}[b]{0.62\textwidth}
  \centering
    \includegraphics[height=0.20\textheight,clip,trim=0 0 0 0]{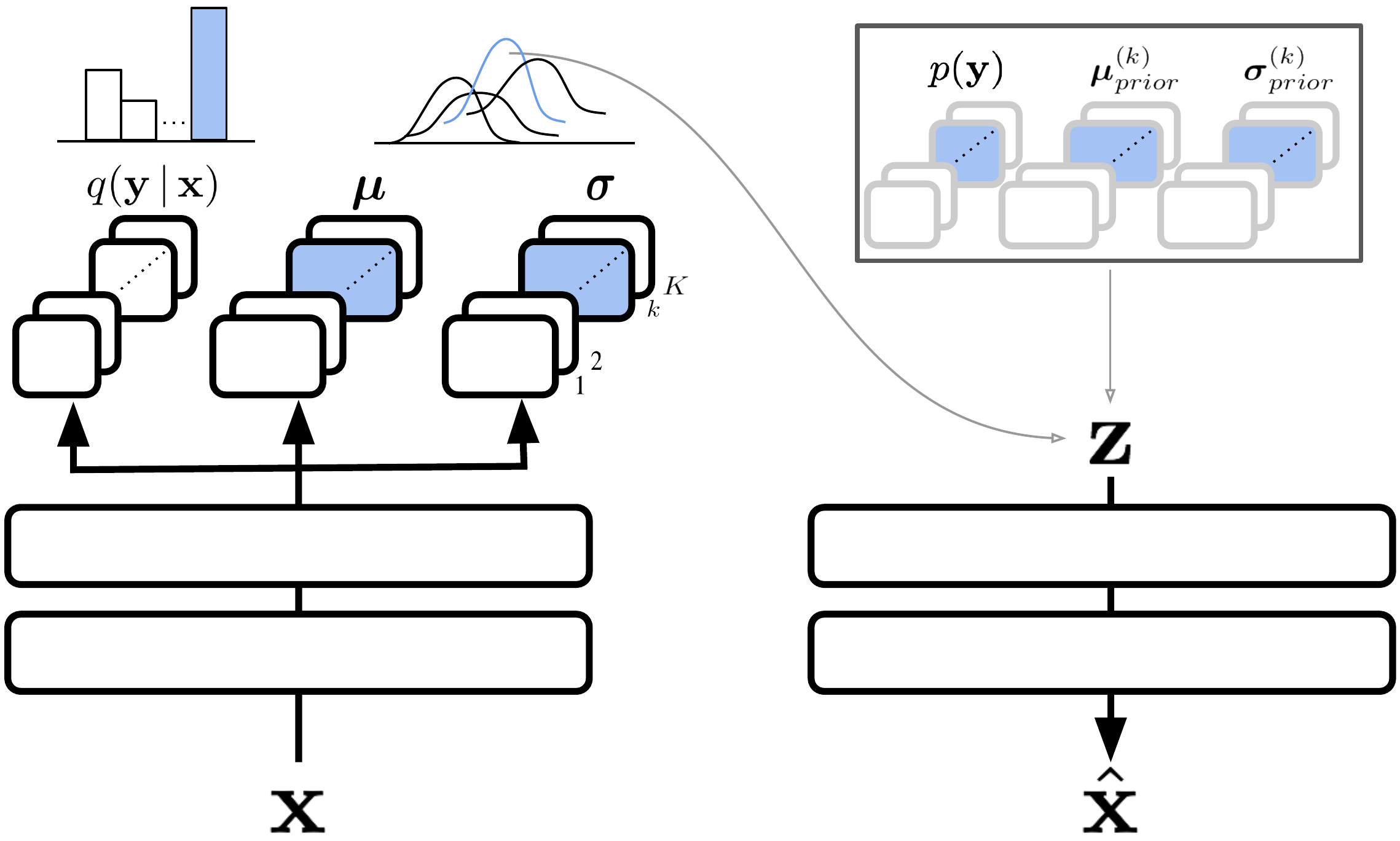}
    \caption{Diagram of the proposed approach, showing the inference procedure and architectural components used.
    }
    \label{fig:network}
\end{minipage}
\end{figure}

\subsection{Inference over tasks}
To address the problem, we utilise the following generative model (Figure~\ref{fig:pgm}):
\begin{eqnarray}
\y &\sim& \textrm{Cat}(\boldsymbol\pi), \nonumber \\
\z &\sim& \mathcal{N}(\boldsymbol\mu_z(\y), \boldsymbol{\sigma}^2_z(\y)), \\
\x &\sim& \textrm{Bernoulli}(\boldsymbol\mu_x(\z)), \nonumber
\end{eqnarray}
\noindent with the joint probability factorising as $p(\x, \y, \z) = p(\y) p(\z \given \y) p(\x \given \z)$. Here, the categorical variable $\y$ indicates the current task, which is then used to instantiate the task-specific Gaussian parameters for latent variable $\z$, which is then decoded to produce the input $\x$.
$p(\y)$ is a fixed uniform prior, with component weights specified by $\bpi$.
In the representation learning scenario, $\y$ can be interpreted as representing some discrete clusters in the data, with $\z$ then representing a mixture of Gaussians which encodes both the inter- and intra-cluster variation.
Posterior inference of $p(\y, \z \given \x)$ in this model is intractable, so we employ an approximate variational posterior $q(\y, \z \given \x) = q(\y \given \x) q(\z \given \x, \y)$.

\noindent Each of these components is parameterised by a neural network: the input is encoded to a shared representation, the mixture probabilities $q(\y \given \x)$ are determined by an output softmax ``task inference'' head, and the Gaussian parameters for $q(\z \given \x, \y=k)$ are produced by the output of a component-specific latent encoding head (one for each component $k$).
The component-specific prior parameters $\boldsymbol\mu_z(\y)$ and $\boldsymbol{\sigma}_z(\y)$ are parameterised as a linear layer (followed by a softplus nonlinearity for the latter) using a one-hot representation of $\y$ as the input.
Finally, the decoder is a single network that maps from the mixture-of-Gaussians latent space $\z$ to the reconstruction $\hat{\x}$.
The architecture is shown in Figure~\ref{fig:network}, where for simplicity, we denote the parameters of the $k^{th}$ Gaussian by $\{ \boldsymbol{\mu}^{(k)}, \boldsymbol{\sigma}^{(k)} \}$.
\noindent The loss for this model is the evidence lower bound (ELBO) given by:
\begin{eqnarray}
\log p(\x) \geq \mathcal{L} & = & \mathbb{E}_{q(\y, \z \given \x)} \left[ \log p(\x, \y, \z) - \log q(\y, \z \given \x) \right] \nonumber \\
                            & = & \mathbb{E}_{q(\y \given \x) q(\z \given \x, \y)} \left[ \log p(\x \given \z) \right] - \mathbb{E}_{q(\y \given \x)} \left[ \kl{q(\z \given \x, \y)}{p(\z \given \y)} \right] \\
                            && - \kl{q(\y \given \x)}{p(\y)} \nonumber
\end{eqnarray}

The expectation over $q(\y \given \x)$ can be computed exactly by marginalising over the $K$ categorical options, but the expectation over $q(\z \given \x, \y)$ is intractable, and requires sampling.
The resulting Monte Carlo approximation comprises a set of familiar terms, some of which correspond clearly to the single-component VAE~\citep{kingma2013auto, rezende2014stochastic}:
\begin{eqnarray}
\mathcal{L} & \approx & \sum_{k=1}^{K}{\overbrace{q(\y = k \given \x)}^{\textrm{component posterior}}\left[ \overbrace{\log p(\x \given \widetilde{\z}^{(k)})}^{\textrm{component-wise reconstruction loss}} - \overbrace{\kl{q(\z \given \x, \y = k)}{p(\z \given \y = k)}}^{\textrm{component-wise regulariser}} \right]} \nonumber \\
            && - \underbrace{\kl{q(\y \given \x)}{p(\y)}}_{\textrm{Categorical regulariser}}
\label{eqn:ELBO}
\end{eqnarray}
\noindent where $\widetilde{\z}^{(k)} \sim q(\z \given \x, \y = k)$ is sampled using the reparametrisation trick.
Of course, this can be generalised to multiple samples in a similar fashion to the Importance-Weighted Autoencoder (IWAE)~\citep{burda2015importance}.

Intuitively, this loss encourages the model to reconstruct the data and perform clustering where possible.
For a given data point, the model can choose to have high entropy over $q(\y \given \x)$, in which case all of the component-wise losses must be low, or assign high $q(\y = k \given \x)$ for some $k$, and use that component to model the datum well.
By exploiting diversity in the input data, the model can learn to utilise different components for different discrete structures (such as classes) in the data.

\subsection{Component-constrained learning}
While our main aim is to operate in an unsupervised setting, there may be cases in which one may wish to train a specific component, or when labels can be generated in a self-supervised fashion.
In such cases where labels $\y_{obs}$ are available, we can use a supervised loss, adapted from Eqn.~\ref{eqn:ELBO}:
\begin{eqnarray}
\mathcal{L}_{sup} & = & \log p(\x \given \widetilde{\z}^{(\y_{obs})}, \y = \y_{obs}) - \kl{q(\z \given \x, \y = \y_{obs})}{p(\z \given  \y = \y_{obs})} \nonumber \\
&& + \log q(\y = \y_{obs} \given \x).
\label{eqn:ELBO_sup}
\end{eqnarray}
\noindent Here, instead of marginalising over $\y$ as in Equation~\ref{eqn:ELBO}, the component-wise ELBO (the first two terms) is computed only for the known label $\y_{obs}$.
Furthermore, the final term in the original ELBO is replaced with a supervised cross-entropy term encouraging $q(\y \given \x)$ to match the label, which reduces to the log posterior probability of the observed label.
This loss will be utilised and further discussed in Sections~\ref{sec:dyn_exp} and \ref{sec:mgr}.

\subsection{Dynamic expansion}
\label{sec:dyn_exp}
To determine the number of mixture components, we opt for a dynamic expansion approach in which capacity is added as needed, by maintaining a small set of poorly-modelled samples and then initialising and fitting a new component to this set when it reaches a critical size.
In a similar fashion to existing techniques such as the Forget-Me-Not process~\citep{milan2016forget} and Dirichlet process~\citep{teh2010dirichlet}, we rely on a threshold to determine when to instantiate a new component.
More concretely, we denote a subset of parameters $\theta^{(k)} = \{ \theta_{q_{y}}^{(k)}, \theta_{q_{z}}^{(k)}, \theta_{p_{z}}^{(k)} \}$ corresponding to the parameters unique to each component $k$ (i.e. the $k^{\textrm{th}}$ softmax output in $q(\y \given \x)$ and the $k^{\textrm{th}}$ Gaussian component in $p(\z \given \y)$ and $q(\z \given \y, \x)$).
During training, any sample with a log-likelihood less than a threshold $\cexp$ is added to set $\Dexp$ (where the log-likelihood is approximated by the ELBO).
Then, when the set $\Dexp$ reaches size $\Nexp$, we initialise the parameters of the new component to the current component $k^{\ast}$ that has greatest probability over $\Dexp$:
\begin{equation}
    \theta^{(K+1)} = \theta^{(k^{\ast})}, \qquad k^{\ast} = \argmax_{k \, \in \{1, 2, \ldots, K \}} \sum_{\x \in  \Dexp} q(\y = k \given \x).
\end{equation}
The new component is then tuned to $\Dexp$, by performing a small fixed number of iterations of gradient descent on all parameters $\theta$, using the component-constrained ELBO (Eqn.~\ref{eqn:ELBO_sup}) with label $K+1$.

Intuitively, this process encourages forward transfer, by initialising new concepts to the ``closest'' existing concept learned by the model and then finetuning to a small number of instances.
The additional capacity used for each expansion is only in the top-most layer of the encoder, with $\sim10^4$ parameters, compared to $\sim2.5\times10^6$ for the rest of the shared model.
That is, while dynamic expansion incorporates a new high-level concept, the underlying low-level representations in the encoder, and the entire decoder, are both shared among all tasks.

\subsection{Combatting forgetting via mixture generative replay}
\label{sec:mgr}
A shared low-level representation can mean that learning new tasks interferes with previous ones, leading to forgetting.
One relevant technique to address this is Deep Generative Replay (DGR)~\citep{shin2017continual}, in which samples from a learned generative model are reused in learning.
We propose to adapt and extend DGR to the mixture setting to perform unsupervised learning without forgetting.
In contrast to the original DGR work, our approach is inherently generative, such that a generative replay-based approach can be incorporated holistically into the framework at minimal cost.
We note that many other existing methods (e.g., ~\citet{kirkpatrick2017overcoming}) could straightforwardly be adapted to our approach, but our experiments demonstrated generative replay to be simple and effective.

To be more precise, during training, the model alternates between batches of real data, with samples $\x_{data}\sim \mathcal{D}$ drawn from the current training distribution, and generated data, with samples $\x_{gen}$ produced by the previous snapshot of the model (with parameters $\tprev$):
\begin{equation}
    \y_{gen} \sim \bpi(\y), \; \z_{gen} \sim \pprev(\z \given \y_{gen}), \; \x_{gen} \sim \pprev(\x \given \z_{gen}),
\end{equation}
\noindent where $\bpi$ represents a choice of prior distribution for the categorical $\y$.
While the uniform prior $p(\y)$ is a natural choice, this fails to consider the degree to which different components are used, and can therefore result in poor sample quality.
To address this, the model maintains a count over components by accumulating the mean of posterior $q(\y \given \x)$ over all previous timesteps, thereby favouring the components that have been used the most.
We refer to this process as \textit{mixture generative replay} (MGR).

While MGR ensures tasks or concepts that have been previously learned by the model are reused for learning, it places no constraint on which components are used to model them.
Given that each generated datum $\x_{gen}$ is conditioned on a sampled $\y_{gen}$, we can use $\y_{gen}$ as a self-supervised learning signal and encourage mixture components to remain consistent with respect to the model snapshot, by using the component-constrained loss from Eqn.~\ref{eqn:ELBO_sup}.

The only remaining question is when to update the previous model snapshot $\tprev$.
For this, we explore two cases, with snapshots taken at periodic \textit{fixed} intervals, or immediately before performing \textit{dynamic} expansion.
The intuition behind the latter is that dynamic expansion is performed when there is a sufficient shift in the input distribution, and consolidating previously learned information is beneficial prior to adding a newly observed concept.
This is also advantageous as it eliminates the additional snapshot period hyperparameter.
\section{Related Work}
\paragraph{Generative models}
A number of related approaches aim to learn a discriminative latent space using generative models.
Building on the original VAE~\citep{kingma2013auto}, \citet{nalisnick2016approximate} utilise a latent mixture of Gaussians, aiming to capture class structure in an unsupervised fashion, and propose a Bayesian non-parametric prior, further developed in~\citep{nalisnick2017stick}.
Similarly,~\citet{joo2019dirichlet} suggest a Dirichlet posterior in latent space to avoid some of the previously observed component-collapsing phenomena.
Lastly,~\citet{jiang2017variational} propose Variational Deep Embedding (VaDE) focused on the goal of clustering in an i.i.d setting.
While VaDE has the same generative process as CURL, it assumes a mean-field approximation, with $\y$ and $\z$ conditionally independent given the input. In the case of CURL, conditioning $\z$ on $\y$ ensures we can adequately capture the inter- and intra- class uncertainty of a sample within the same structured latent space $\z$.

\paragraph{Continual learning}
A large body of work has addressed the continual learning problem~\citep{parisi2019continual}.
\textit{Regularisation-based} methods minimise changes to parameters that are crucial for earlier tasks, with some parameter-wise weight to measure importance~\citep{kirkpatrick2017overcoming, nguyen2017variational, zenke2017continual, aljundi2018memory, schwarz2018progress}. Related techniques seek to ensure the performance on previous data does not decrease, by employing constrained optimisation~\citep{lopez2017gradient, chaudhry2018efficient} or distilling the information from old models or tasks~\citep{li2018learning}. 
In a similar vein, other methods encourage new tasks to utilise previously unused parameters, either by finding ``free'' linear parameter subspaces~\citep{he2018overcoming}; learning an attention mask over parameters~\citep{serra2018overcoming}; or using an agent to find new activation paths through a network~\citep{fernando2017pathnet}.
\textit{Expansion-based} models dynamically increase capacity to allow for additional tasks~\citep{rusu2016progressive, yoon2017lifelong, draelos2017neurogenesis}, and optionally prune the network to constrain capacity~\citep{zhou2012online, golkar2019continual}.
Another popular approach is that of \textit{rehearsal-based} methods~\citep{robins1995catastrophic}, where the data distribution from earlier tasks is captured by samples from a generative model trained concurrently~\citep{shin2017continual, van2018generative, ostapenko2018learning}.
\citet{farquhar2018towards} combine such methods with regularisation-based approaches under a Bayesian interpretation.
Alternatively,~\citet{rebuffi2017icarl} learn class-specific exemplars instead of a generative model.
However, these methods usually require task identities, rely on well-defined task boundaries, and are often evaluated on a sequence of supervised learning tasks.

\paragraph{Task-agnostic continual learning}
Some recent work has investigated continual learning without task labels or boundaries.
\citet{hsu2018re} and \citet{van2019three} identify the scenarios of incremental task, domain, and class learning; which operate without task labels in the latter cases, but all focus on supervised learning tasks.
\citet{aljundi2019task} propose a task-free approach to continual learning related to ours, which mitigates forgetting using the regularisation-based Memory Aware Synapses (MAS) approach~\citep{aljundi2018memory}, maintains a hard example buffer to better estimate the regularisation weights, and detects when to update these weights (usually performed at known task boundaries in previous work).
\citet{zeno2018task} propose a Bayesian task-agnostic learning update rule for the mean and variance of each parameter, and demonstrate its ability to handle ambiguous task boundaries.
However, it is only applied to supervised tasks, and can exploit the ``label'' trick, inferring the task based on the class label.
In contrast,~\citet{achille2018life} address the problem of unsupervised learning in a sequential setting by learning a disentangled latent space with task-specific attention masks, but the main focus is on learning across datasets, and the method relies on abrupt shifts in data distribution between datasets.
Our approach builds upon this existing body of work, addressing the full unsupervised continual learning problem, where task labels and boundaries are unknown, and the tasks themselves are without class supervision.
We argue that addressing this problem is critical in order to tackle continual learning in challenging, real-world scenarios.
\section{Experiments}

In the following sections, we empirically evaluate a) whether our method learns a meaningful class-discriminable latent space in the unsupervised sequential learning setting, without forgetting, even when task boundaries are unclear; b) the importance of the dynamic expansion and generative replay techniques to performance; and c) how CURL performs on external benchmarks when trained i.i.d or adapted to learn in a supervised fashion.
Code for all experiments can be found at \url{https://github.com/deepmind/deepmind-research/}.

%%%%%%%%%%%%%%%%%%%%%%%%%%%%%%%%%%%%%%%%%%%%%%

\subsection{Evaluation settings and datasets}
One desired outcome of our approach is the ability to learn class-discriminative latent representations from non-stationary input data. We evaluate this using cluster accuracy (the accuracy obtained when assigning each mixture component to its most represented class), and with the accuracy of a k-Nearest Neighbours (k-NN) classifier in latent space.
The former measures the amount of class-relevant information encoded into the categorical variable $\y$, while the latter measures the discriminability of the entire latent space without imposing structure (such as a linear boundary).

For the evaluation we extensively utilise the MNIST~\citep{lecun2010mnist} and Omniglot~\citep{lake2011one} datasets, and further information can be found in Appendix~\ref{sec:datasets}.
We investigate a number of different evaluation settings: \textit{i.i.d}, where the model sees shuffled training data; \textit{sequential}, where the model sees classes sequentially; and \textit{continuous drift}, similar to the sequential case, but with classes gradually introduced by slowly increasing the number of samples from the new class within a batch.

%%%%%%%%%%%%%%%%%%%%%%%%%%%%%%%%%%%%%%%%%%%%%%

\subsection{Continual class-discriminative representation learning}
\label{sec:cl}
We begin by analysing our approach, and follow this with evaluation on external benchmarks in later sections.
First, we measure the ability to perform class-discriminative representation learning in the \emph{sequential} setting on MNIST, where each of the classes is observed for $10000$ training steps (further experimental details can be found in Appendix~\ref{sec:main_setup}).
Figure~\ref{fig:cl_plots}a shows the cluster accuracy for a number of variants of CURL.
We observe the importance of both dynamic expansion and mixture generative replay (MGR) to learn a coherent representation without forgetting.
Figure~\ref{fig:cl_plots}b shows the class-wise accuracies during training, for the model with MGR and expansion.
Interestingly, while many existing continual learning approaches appear to forget earlier classes (see e.g.~\citet{nguyen2017variational}), these classes are well modelled by CURL, and the confusion is more observed between similar classes (such as $3$s and $5$s; or $7$s and $9$s).
Indeed, this is reflected in the class-confusion matrix after training (Figure~\ref{fig:cl_plots}c).
This implies the model adequately addresses catastrophic forgetting, but could improve in terms of plasticity, i.e., learning new concepts.
Further analysis can be found in Appendix~\ref{sec:gen_samples}, showing generated samples; and Appendix~\ref{sec:data_buffers}, analysing the dynamic expansion buffers.

\begin{figure}[t]
\begin{subfigure}[t]{0.25\textwidth}
  \centering
    \includegraphics[scale = 0.28, clip, trim=0cm 0cm 0 0.2cm]{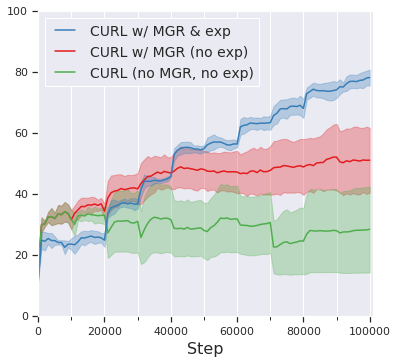}
    \caption{}
\end{subfigure}
\hspace{0.5cm}
\begin{subfigure}[t]{0.4\textwidth}
  \centering
    \includegraphics[scale = 0.25, clip,trim=2cm 1.2cm 1.5cm 0]{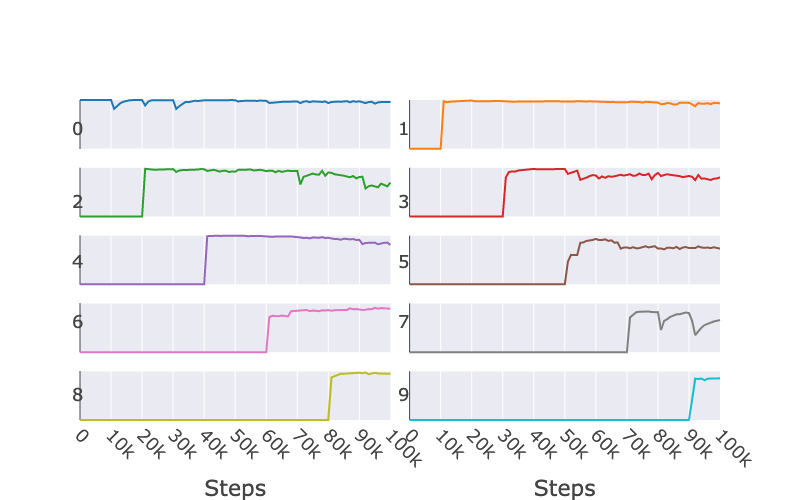}
    \caption{}
\end{subfigure}
\hspace{0.5cm}
\begin{subfigure}[t]{0.25\textwidth}
  \centering
    \includegraphics[scale = 0.23, clip,trim=0 0 2cm 0]{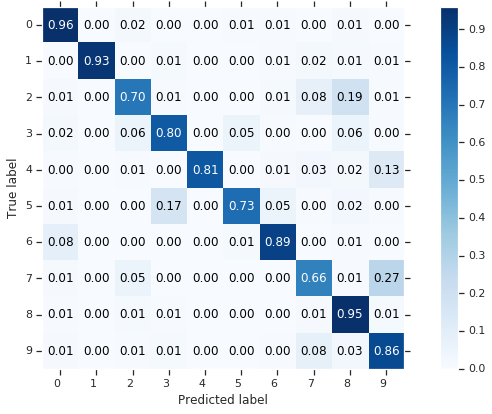}
    \caption{}
\end{subfigure}
\caption{a) Cluster accuracy for CURL variants on MNIST, measuring the contribution of mixture generative replay (``MGR'') and dynamic expansion (``exp''); b) Accuracy per class, over time; c) Class confusion matrix at the end of learning, for CURL w/ MGR \& exp.}
\label{fig:exp_ablation}
\end{figure}

%%%%%%%%%%%%%%%%%%%%%%%%%%%%%%%%%%%%%%%%%%%%%%

\subsection{Ablation studies}\label{sec:ablation}
Next, we perform an ablation study to gauge the impact of the expansion threshold for continual learning, in terms of cluster accuracy and number of components used, as shown in Figure~\ref{fig:exp_ablation}.
As the threshold value is increased, samples are more frequently stored into the ``poorly-modelled'' buffer, and the model expands more aggressively throughout learning.
Consequently, for sequential learning, the number of components ranges from $12$ to $71$, the cluster accuracy varies up to a maximum of $84\%$, and the $k$-NN error also marginally decreases over this range.
Furthermore, without any dynamic expansion, the result is significantly poorer at $51\%$ accuracy, and when discovering the same number of components with dynamic expansion ($25$, obtained with an expansion threshold of $-200$), the equivalent performance is at $77\%$.
Thus, the dynamic expansion threshold conveniently provides a tuning parameter to perform capacity estimation, trading off cluster accuracy with the memory cost of using additional components in the latent mixture.
Interestingly, if we perform the same analysis for i.i.d. data (also in Figure~\ref{fig:exp_ablation}), we observe a similar trade-off; though the final performance is slightly poorer than when starting with an equivalent, fixed number of mixture components ($22$).

We also further analyse mixture generative replay (MGR) with an ablation study in Table~\ref{table_mgr_ablation}.
We evaluate standard and self-supervised MGR (SMGR), and compare between the case where snapshots are taken on expansion (i.e., no task information is needed), or at fixed intervals (either at $T$, the duration of training on each class, or $0.1T$, ten times more frequently).
Intuitively, the period is important as it determines how quickly a shifting data distribution is consolidated into the model: if too short, the generated data will drift with the model, leading to forgetting.
The results in Table~\ref{table_mgr_ablation} point to a number of interesting observations.
First, both MGR and SMGR are sensitive to the fixed snapshot period: the performance is unsurprisingly optimal when snapshots are taken as the training class changes, but drops significantly when performed more frequently, and also uses a greater number of clusters in the process.
Second, by taking snapshots before dynamic expansion instead, this performance can largely be recovered, and without any knowledge of the task boundaries.
Third, perhaps surprisingly, SMGR harms performance compared to MGR.
This may be due to the fact that mixture components already tend to be consistent in latent space throughout learning, and SMGR may be reducing plasticity; further analysis can be found in Appendix~\ref{sec:latent_space}.
Lastly, we can also observe the benefits of MGR, with the MNIST case exhibiting far poorer performance and utilising many more components in the process.
Interestingly, the Omniglot case without MGR performs well, but at the cost of significantly more components: expansion itself is able to partly address catastrophic forgetting by effectively oversegmenting the data.

\begin{figure}[t]
    \begin{subfigure}[t]{\linewidth}
        \includegraphics[scale = 0.35, clip, trim=0 0 0 0]{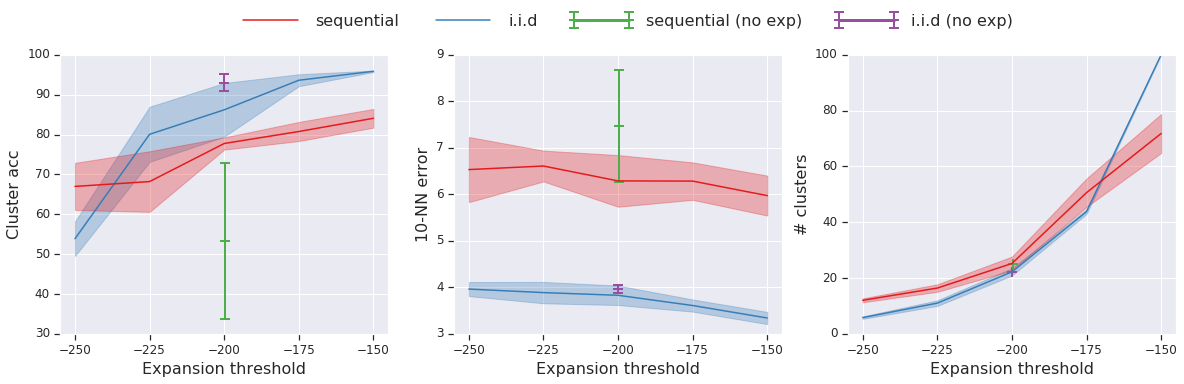}
        \begin{tabularx}{\textwidth}{*{3}{>{\centering\arraybackslash}X}}
        \caption{}
        & \caption{}
        & \caption{}
        \end{tabularx}
    \end{subfigure}
    \vspace*{-5mm}
\caption{Ablation study for dynamic expansion on MNIST, showing (a) cluster accuracy; (b) $10$-NN error; and (c) number of components used; when varying the expansion threshold $c_{exp}$. For comparison, we also show the performance without expansion (``no exp''), but using the same number of components as in the $c_{exp} = -200$ case.}
\label{fig:cl_plots}
\end{figure}

\begin{table}[t]
\centering
\scalebox{0.7}{
\begin{tabular}{ c | c c c | c c c }
  \hline \\ [-1.5ex]
 \bf{Benchmark} & \multicolumn{3}{c}{\bf{MNIST}} & \multicolumn{3}{c}{\bf{Omniglot}} \\
 Scenario & \# clusters & Cluster acc (\%) $\uparrow$ & $10$-NN error (\%) $\downarrow$ & \# clusters & Cluster acc (\%) $\uparrow$ & $10$-NN error (\%) $\downarrow$  \\
 \hline \\ [-1.5ex]
 MGR (fixed, T) & $25.20_{\pm 2.23}$ & $77.74_{\pm 1.37}$ & $6.29_{\pm 0.50}$ & $101.20_{\pm 8.45}$ & $13.21_{\pm 0.53}$ & $76.34_{\pm 1.10}$ \\
 MGR (fixed, 0.1T) & $37.60_{\pm 2.15}$ & $49.14_{\pm 3.95}$ & $14.95_{\pm 0.73}$ & $131.60_{\pm 15.74}$ & $12.13_{\pm 1.54}$ & $81.21_{\pm 2.06}$ \\
 MGR (dyn) & $35.20_{\pm 2.79}$ & $57.76_{\pm 1.43}$ & $12.08_{\pm 1.19}$ & $127.20_{\pm 16.67}$ & $12.74_{\pm 0.60}$ & $80.56_{\pm 1.39}$ \\
 SMGR (fixed, T) & $28.20_{\pm 0.40}$ & $69.27_{\pm 1.46}$ & $7.50_{\pm 0.57}$ & $105.20_{\pm 5.56}$ & $11.32_{\pm 0.52}$ & $76.62_{\pm 1.49}$ \\
 SMGR (fixed, 0.1T) & $39.80_{\pm 6.05}$ & $48.18_{\pm 1.72}$ & $15.48_{\pm 0.81}$ & $137.40_{\pm 9.75}$ & $9.01_{\pm 2.17}$ & $85.73_{\pm 5.84}$ \\
 SMGR (dyn) & $36.00_{\pm 2.45}$ & $53.97_{\pm 3.52}$ & $11.72_{\pm 1.16}$ & $152.20_{\pm 25.02}$ & $10.48_{\pm 1.10}$ & $84.44_{\pm 4.10}$ \\
 \hline \\ [-1.5ex]
 CURL (no MGR) & $55.80_{\pm 1.94}$ & $45.35_{\pm 1.50}$ & $17.46_{\pm 1.25}$ & $189.60_{\pm 9.75}$ & $13.36_{\pm 1.06}$ & $81.91_{\pm 1.36}$ \\
 \hline
\end{tabular}}
\vspace{1ex}
\caption{Ablation study for mixture generative replay (MGR and SMGR), indicating the performance and number of components used. All variants perform dynamic expansion}
\label{table_mgr_ablation}
\end{table}

%%%%%%%%%%%%%%%%%%%%%%%%%%%%%%%%%%%%%%%%%%%%%%

\subsection{Learning with poorly-defined task boundaries}
\begin{figure}[t]
\begin{minipage}[b]{0.55\textwidth}
  \centering%
  \includegraphics[scale=0.15,clip,trim=1.5cm 1.2cm 0 3cm]{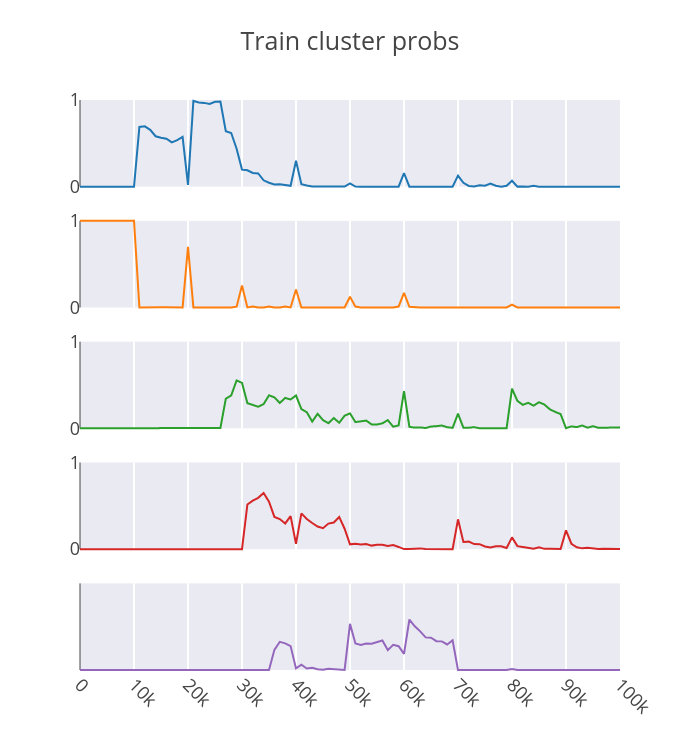}
  \includegraphics[scale=0.15,clip,trim=1.5cm 1.2cm 0 3cm]{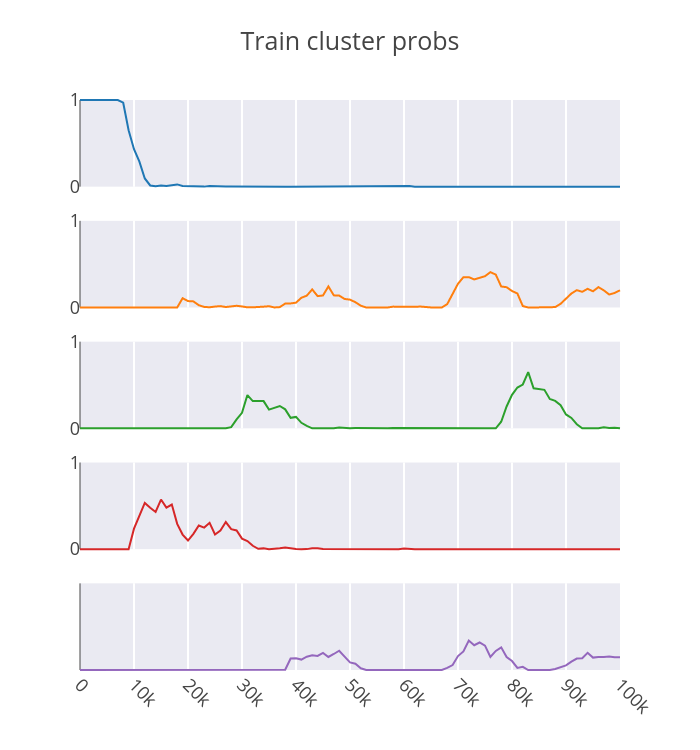}
  \captionof{figure}{Mixture probabilities of the $5$ components used most throughout training, with discrete class changes (left), and with continuous class drift (right).}
  \label{fig:continuous}
\end{minipage}
\hfill
\begin{minipage}[b]{0.4\textwidth}
\centering%
\scalebox{0.8}{
     \begin{tabular}{ c | c c }
     \hline \\ [-1.5ex]
     \bf{Benchmark} & \multicolumn{2}{c}{\bf{SplitMNIST}} \\
     Evaluation & Incr. Task & Incr. Class \\
     \hline \\ [-1.5ex]
     EWC & $98.64_{\pm0.22}$ & $20.01_{\pm0.06}$ \\
     %Online EWC & $99.12_{\pm0.11}$ & $19.96_{\pm0.07}$ \\
     SI & $99.09_{\pm0.15}$ & $19.99_{\pm0.06}$ \\
     MAS & $99.22_{\pm0.21}$ & $19.52_{\pm0.29}$ \\
     LwF & $\bold{99.60_{\pm0.03}}$ & $24.17_{\pm0.33}$ \\
     GEM & $98.42_{\pm0.10}$ & $92.20_{\pm0.12}$ \\
     DGR & $99.50_{\pm0.03}$ & $91.24_{\pm0.33}$ \\
     iCARL & - & $\bold{94.57_{\pm0.11}}$ \\
     CURL & $99.10_{\pm 0.06}$ & $92.59_{\pm0.66}$ \\
     \hline
    \end{tabular}}
    \captionof{table}[]{Supervised learning benchmark on splitMNIST, for incremental task and incremental class learning.{\footnotemark}}
    \label{table:supervised}
\end{minipage}
\end{figure}

Next, we evaluate CURL in the \textit{continuous drift} setting, and compare to the standard sequential setting.
The overall performance on MNIST and Omniglot is shown in Table~\ref{table:continuous}, using MGR with either fixed or dynamic snapshots.
We observe that despite having unclear task boundaries, where classes are gradually introduced, the continuous case generally exhibits better performance than the case with well-defined task boundaries.
We also closely investigate the mixture component dynamics during learning, by obtaining the top $5$ components (most used over the course of learning) and plotting their posterior probabilities over time (Figure~\ref{fig:continuous}).
From the discrete task-change domain (left), we observe that probabilities change sharply with the hard task boundaries (every $10000$ steps); and many mixture components are quite sparsely activated, modelling either a single class, or a few classes.
Some of the mixture components also observe ``echoes'', where the sharp change to a new class in the data distribution activates the component temporarily before dynamic expansion is performed.
In the continuous drift case (right of Figure~\ref{fig:continuous}), the mixture probabilities exhibit similar behaviours, but are much smoother in response to the gradually changing data distribution. Further, without a sharp distributional shift, the ``echoes'' are not observed.

\begin{table}[t]
\centering
\scalebox{0.7}{
\begin{tabular}{ c | c c c | c c c }
  \hline \\ [-1.5ex]
 \bf{Benchmark} & \multicolumn{3}{c}{\bf{MNIST}} & \multicolumn{3}{c}{\bf{Omniglot}} \\
 Scenario & \# clusters & Cluster acc (\%) $\uparrow$ & $10$-NN error (\%) $\downarrow$ & \# clusters & Cluster acc (\%) $\uparrow$ & $10$-NN error (\%) $\downarrow$ \\
 \hline \\ [-1.5ex]
 Seq. w/ MGR (fixed) & $25.20_{\pm 2.23}$ & $77.74_{\pm 1.37}$ & $6.29_{\pm 0.50}$ & $101.20_{\pm 8.45}$ & $13.21_{\pm 0.53}$ & $76.34_{\pm 1.10}$ \\
 Seq. w/ MGR (dyn) & $35.20_{\pm 2.79}$ & $57.76_{\pm 1.43}$ & $12.08_{\pm 1.19}$ & $127.20_{\pm 16.67}$ & $12.74_{\pm 0.60}$ & $80.56_{\pm 1.39}$ \\
 \hline \\ [-1.5ex]
 Cont. w/ MGR (fixed) & $44.60_{\pm 2.65}$ & $79.38_{\pm 4.26}$ & $6.56_{\pm 0.42}$ & $111.40_{\pm 3.77}$ & $13.17_{\pm 0.37}$ & $75.80_{\pm 1.19}$ \\
 Cont. w/ MGR (dyn) & $50.40_{\pm 1.85}$ & $64.93_{\pm 2.09}$ & $9.88_{\pm 1.43}$ & $129.20_{\pm 2.14}$ & $13.54_{\pm 0.35}$ & $78.78_{\pm 0.39}$ \\
 \hline
\end{tabular}}
\vspace{1ex}
\caption{Performance comparison between the sequential learning setting (with discrete changes in class), versus the continuous drift setting (with class ratios gradually changing).}
\label{table:continuous}
\end{table}

\footnotetext{Performances of existing approaches are taken from studies by~\citet{hsu2018re} and~\citet{van2019three}, using the better of the two.}
%%%%%%%%%%%%%%%%%%%%%%%%%%%%%%%%%%%%%%%%%%%%%%

\subsection{External benchmarks}

\paragraph{Supervised continual learning}
While focused on task-agnostic continual learning in unsupervised settings, CURL can also be trivially adapted to supervised tasks simply by training with the supervised loss in Eqn.~\ref{eqn:ELBO_sup}.
We evaluate on the split MNIST benchmark, where the data are split into five tasks, each classifying between two classes, and the model is trained on each task sequentially.
If we evaluate the overall accuracy after training, this is called \textit{incremental class learning}; and if we provide the model with the appropriate task label and evaluate the binary classification accuracy for each task, this is \textit{incremental task learning}~\citep{hsu2018re, van2019three}.
Experimental details can be found in Appendix~\ref{sec:supervised_setup}.
The results in Table~\ref{table:supervised} demonstrate that the proposed unsupervised approach can easily and effectively be adapted to supervised tasks, achieving competitive results for both scenarios.
While all methods perform quite well on incremental task learning, CURL is outperformed only by iCARL~\citep{rebuffi2017icarl} on incremental class learning, which was specifically proposed for this task.
Interestingly, the result is also better than DGR, suggesting that by holistically incorporating the generative process and classifier into the same model, and focusing on the broader unsupervised, task-agnostic perspective, CURL is still effective in the supervised domain.

\begin{table}[t]
\centering
\scalebox{0.8}{
\begin{tabular}{ c | c c c | c c c } 
  \hline \\ [-1.5ex]
 \bf{Benchmark} & \multicolumn{3}{c}{\bf{MNIST ($n_z$ = 50)}} & \multicolumn{3}{c}{\bf{Omniglot ($n_z$ = 100)}} \\
 Evaluation & $3$-NN error & $5$-NN error & $10$-NN error & $3$-NN error & $5$-NN error & $10$-NN error  \\
 \hline \\ [-1.5ex]
 VAE\footnotemark & $27.16_{\pm0.48}$ & $20.20_{\pm0.93}$ & $14.89_{\pm0.40}$ & $92.34_{\pm0.25}$ & $91.21_{\pm0.18}$ & $88.79_{\pm0.35}$ \\
 SBVAE\footnotemark[\value{footnote}] & $10.01_{\pm0.52}$ & $9.58_{\pm0.47}$ & $9.39_{\pm0.54}$ & $86.90_{\pm0.82}$ & $85.10_{\pm0.89}$ & $82.96_{\pm0.64}$ \\
 DirVAE\footnotemark[\value{footnote}] & $5.98_{\pm0.06}$ & $5.29_{\pm0.06}$ & $5.06_{\pm0.06}$ & $\bold{76.55_{\pm0.23}}$ & $\bold{73.81_{\pm0.29}}$ & $\bold{70.95_{\pm0.29}}$ \\
 CURL (i.i.d) & $\bold{4.40_{\pm 0.34}}$ & $\bold{4.22_{\pm 0.28}}$ & $\bold{4.23_{\pm 0.30}}$ & $78.18_{\pm 0.47}$ & $75.41_{\pm 0.34}$ & $72.51_{\pm 0.46}$ \\
 VaDE (bigger net) & $\bold{2.20}$ & $\bold{2.14}$ & $\bold{2.22}$ & - & - & - \\
 \hline \\ [-1.5ex]
 CURL w/ MGR (seq) & $4.58_{\pm 0.26}$ & $4.35_{\pm 0.32}$ & $4.50_{\pm 0.34}$ & $83.95_{\pm 0.72}$ & $81.56_{\pm 0.75}$ & $78.80_{\pm 0.74}$ \\
 \hline \\ [-1.5ex]
 Raw pixels\footnotemark[\value{footnote}] & $3.00$ & $3.21$ & $3.44$ & $69.94$ & $69.41$ & $70.10$ \\
 \hline
\end{tabular}}
\vspace{1ex}
\caption{Unsupervised learning benchmark comparison with sampled latents. We compare with a number of approaches trained i.i.d, as well as CURL trained in the sequential setting.}
\label{table:perf}
\end{table}
\footnotetext{Performance numbers are obtained from \citet{joo2019dirichlet}, with consistent architectures and hyperparameters.}

\paragraph{Unsupervised i.i.d learning}
We also demonstrate the ability of the underlying model to learn in a more traditional setting with the entire dataset shuffled, and compare with existing work in clustering and representation learning: the VAE~\citep{kingma2013auto}, DirichletVAE~\citep{joo2019dirichlet}, SBVAE~\citep{nalisnick2017stick}, and VaDE~\citep{jiang2017variational}.
We utilise the same architecture and hyperparameter settings as in~\citet{joo2019dirichlet} for consistency, with latent spaces of dimension $50$ and $100$ for MNIST and Omniglot respectively; and full details of the experimental setup can be found in Appendix~\ref{sec:clustering_setup}.
We note that the $k$-NN error values are much better here than in Section~\ref{sec:ablation}; this is due to a higher dimensional latent space and hence they cannot be directly compared (see Appendix~\ref{sec:knn_dim}).

The uppermost group in Table~\ref{table:perf} show the results on i.i.d MNIST and Omniglot.
The CURL generative model trained i.i.d (without MGR, and with dynamic expansion) is competitive with the state-of-the-art on MNIST (bettered only by VaDE, which incorporates a larger architecture) and Omniglot (bettered only by DirVAE).
While not the main focus of this paper, this demonstrates the ability of the proposed generative model to learn a structured, discriminable latent space, even in more standard learning settings with shuffled data.
Table~\ref{table:perf} also shows the performance of CURL trained in the \textit{sequential} setting.
We observe that, despite learning from sequential data, these results are competitive with the state-of-the-art approaches that operate on i.i.d. data.

\section{Conclusions}
In this work, we introduced an approach to address the unsupervised continual learning problem, in which task labels and boundaries are unknown, and the tasks themselves lack class labels or other external supervision.
Our approach, named CURL, performs task inference via a mixture-of-Gaussians latent space, and uses dynamic expansion and mixture generative replay (MGR) to instantiate new concepts and minimise catastrophic forgetting.
Experiments on MNIST and Omniglot showed that CURL was able to learn meaningful class-discriminative representations without forgetting in a sequential class setting (even with poorly defined task boundaries).
External benchmarks also demonstrated the method to be competitive with respect to previous work when adapted to unsupervised learning from i.i.d data, and to supervised incremental class learning.
Future directions will investigate additional techniques to alleviate forgetting, and the extension to the reinforcement learning domain.

\bibliography{references}
\bibliographystyle{bibstyle}

\appendix
\newpage
{\bf{\LARGE{Appendix}}}
\section{Additional experiments}

\subsection{Generated samples}
\label{sec:gen_samples}
The primary aim of this paper is to learn a meaningful class-discriminative representation of the data.
We use mixture generative replay to contrast catastrophic forgetting, hence sample quality is not our main interest.
Nonetheless, we are interested in checking whether the model is able to capture the variety of the data and what level of sample quality is sufficient to retain what has been learned.
We illustrate samples generated after sequentially learning on each class, in Figure~\ref{fig:gen_samples}.
We observe from the later samples (bottom rows) that classes that are observed early on are still preserved within the model.
Interestingly, while most of the samples are clear, some indicate degraded but still identifiable versions of previous symbols (such as some of the zeros after other classes are introduced).
Given that the learned latent representations are still class-discriminative, we hypothesise that perhaps merely capturing the ``essence'' of a particular class may be more useful for representation learning than striving for a pixel-perfect reconstruction.
We leave a thorough analysis of this idea to future work.

\subsection{Data buffers for dynamic expansion}
\label{sec:data_buffers}
\begin{figure}[h]
\begin{minipage}{0.45\textwidth}
  \centering
    \includegraphics[scale = 0.15, clip,trim=0 0 0 0]{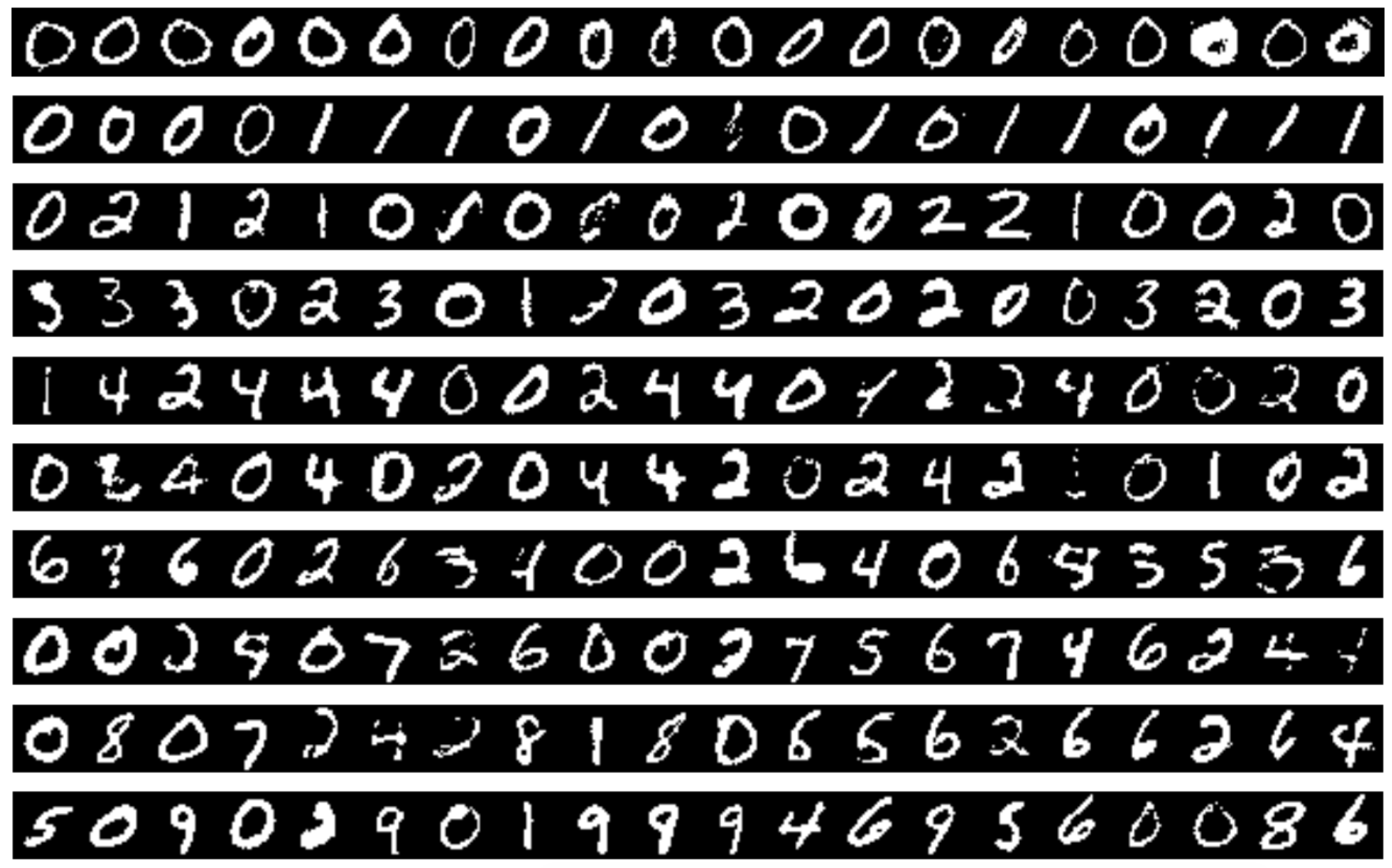}
    \caption{Model-generated samples, with each row corresponding to observing the next class in the sequential setting (top to bottom).} 
  \label{fig:gen_samples}
\end{minipage}
\hspace{0.5cm}
\begin{minipage}{0.5\textwidth}
  \centering
    \includegraphics[scale = 0.17, clip,trim=0 17cm 0 0]{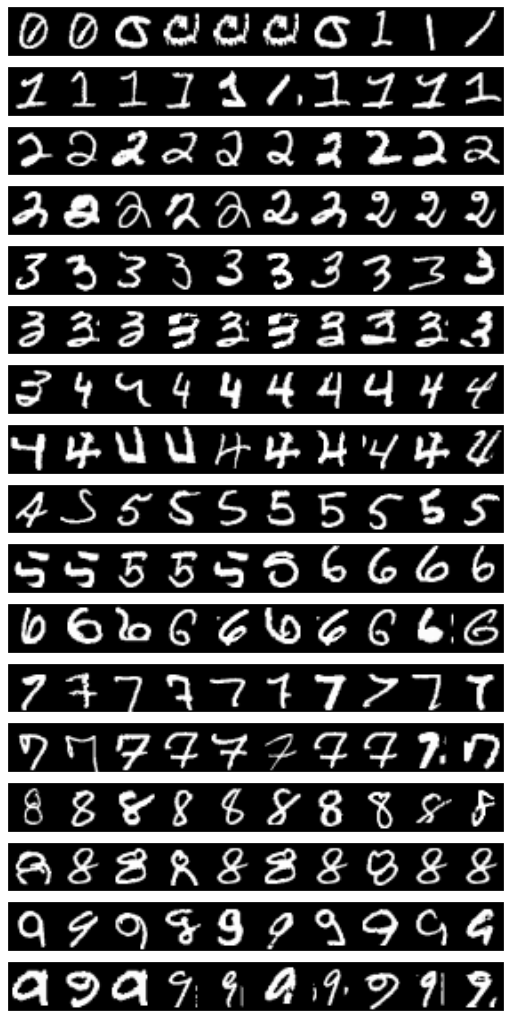}
    \quad
    \includegraphics[scale = 0.17, clip,trim=0 -2cm 0 19cm]{poor_buffer_samples.png}
    \caption{Samples stored in the poor data buffer before each successive expansion (top to bottom; left to right).}
    \label{fig:poor_buffer}
\end{minipage}
\end{figure}

We can also observe samples from the data buffers used for each dynamic expansion, to understand when the model chooses to expand capacity (Figure~\ref{fig:poor_buffer}).
We note that some buffers holds samples from multiple classes: this occurs when there are few outliers from a single class, insufficient to initialise a new component on their own.
Nonetheless, we observe an intuitive trend over most of the buffers: in many cases for a class, the first expansion denotes the change in distribution that corresponds to the introduction of the class; and the second expansion responds to outlying or challenging examples from the class.
This can be observed, for instance, for the initialisation of a component for ``twos'' and then for ``curly twos'', or for ``threes'' and ``challenging threes''.
This highlights the ability of the approach to incorporate new components to account for both data distributional shift and hard example modelling.

%%%%%%%%%%%%%%%%%%%%%%%%%%%%%%%%%%%%%%%%%%%%%%

\subsection{Latent structure}
\label{sec:latent_space}

\begin{figure}[h!]
\begin{center}
\begin{subfigure}{\textwidth}
  \includegraphics[scale=0.23,clip,trim=0 24.5cm 0 0]{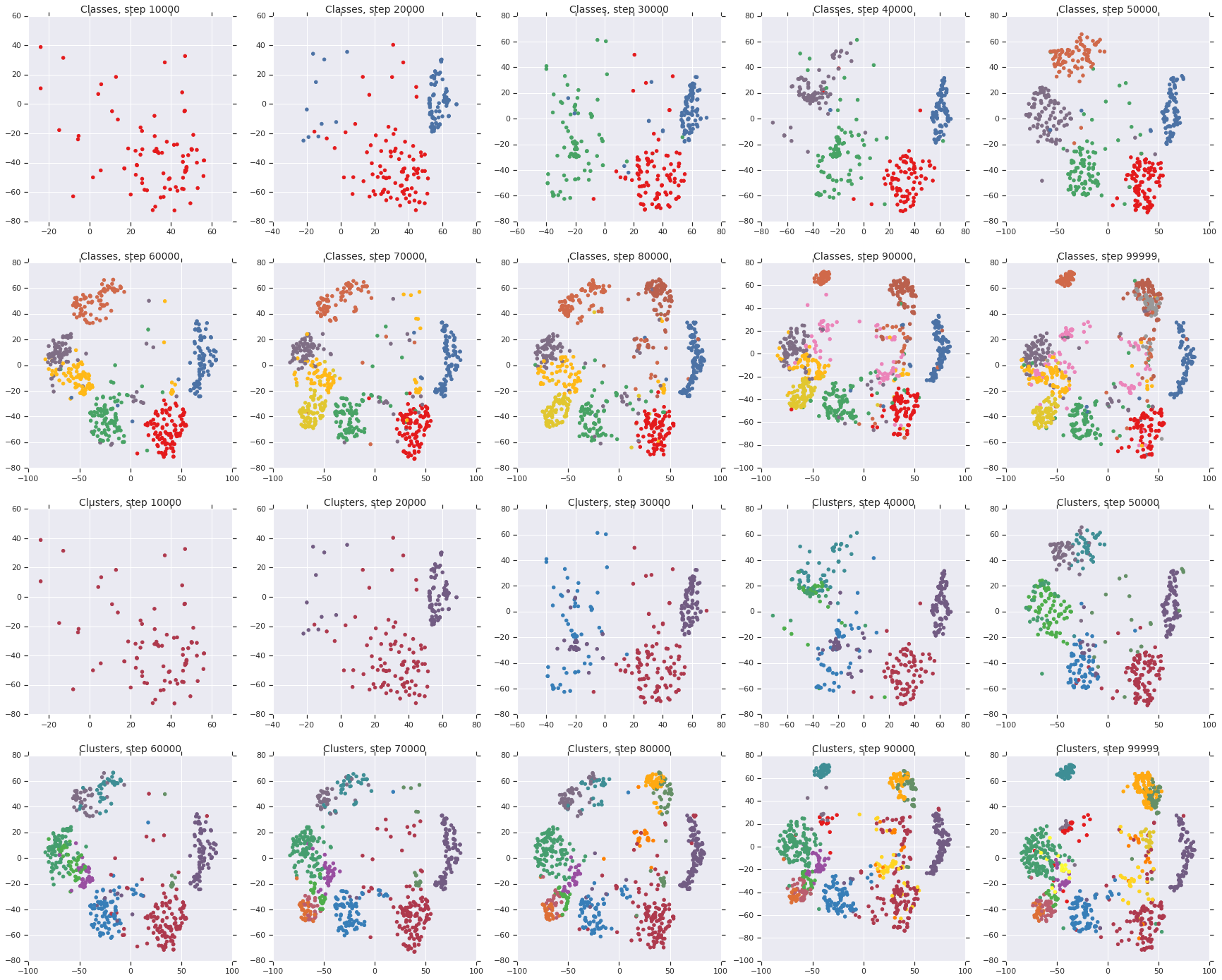}
  \caption{}
\end{subfigure}
\begin{subfigure}{\textwidth}
  \includegraphics[scale=0.23,clip,trim=0 0 0 24.5cm]{latent_space.png}
  \caption{}
\end{subfigure}
\end{center}
\caption{t-SNE plots showing the evolution of the latent space during learning, with points coloured by (a) class label, and (b) mixture component.}
\label{fig:tsne}
\end{figure}

To analyse the structure of our latent space, we encode our test set to latent space, and observe how the latent space changes during training (Figure~\ref{fig:tsne}).
Figure~\ref{fig:tsne}(a) colours the points by ground truth class label (not available to the model during learning), and Figure~\ref{fig:tsne}(b) colours them by the most probable mixture component (i.e. $\argmax_k q(\y = k \given \x)$).
Each subplot represents a particular step during training, and only points from classes seen so far are used in the plot.
We observe from Figure~\ref{fig:tsne}(a) that as classes are incrementally introduced, they occupy a relatively disjoint region of latent space, and are consistent over time.
That is, the introduction of new classes does not appear to catastrophically interfere with the learned representations of previous ones.

We also observe similar properties in Figure~\ref{fig:tsne}(b), with the clusters covering individual classes with reasonable accuracy, and maintaining a consistent position over time.
This may offer some explanation for the inefficacy of SMGR - there is little value to incorporating the self-supervised loss for component consistency given that the mixture components are already consistent in which classes they model.

%%%%%%%%%%%%%%%%%%%%%%%%%%%%%%%%%%%%%%%%%%%%%%

\subsection{Discussion on k-NN accuracy measure}
\label{sec:knn_dim}
We exploit the $k$-NN accuracy in our experiments to measure the class-discriminability of latent space, without imposing any specific parametric structure in terms of the boundary between classes.
As a simple non-parametric approach, it serves this purpose, and has also been used extensively in previous evaluations as a result~\citep{nalisnick2017stick, joo2019dirichlet}.
However, there are some interesting properties that are worth considering.
Firstly, as demonstrated by Table~\ref{table:knn_dim}, the measure is highly dependent on dimensionality, making it difficult to compare across different latent space sizes.
In our experiments, all architectural aspects, including latent size, are kept fixed for an experiment to account for this.
Secondly, Simple baselines like classifying on raw pixels perform surprisingly well (e.g., approximately $3\%$ k-NN error on MNIST); this is due in part to the much larger dimensionality than the latent spaces used for evaluation, but we postulate that this is also likely due to the image statistics of the datasets used: given MNIST and Omniglot have many low-variance black pixels and a relatively small amount of intra-class variance within the centre pixels, the raw pixels are themselves discriminative.
Given the inability to directly compare between different dimensionalities, this simple baseline is only provided for context.

\begin{table}[h!]
\centering
\scalebox{0.75}{
\begin{tabular}{ c | c c c c c c c c }
  \hline \\ [-1.5ex]
  & \multicolumn{8}{c}{\bf{Latent space dimension} ($n_z$)} \\
  & 8 & 16 & 32 & 64 & 128 & 256 & 784 & 1568 \\
  \hline \\ [-1.5ex]
  $10$-NN error & $58.75_{\pm2.94}$ & $42.12_{\pm1.76}$ & $28.20_{\pm1.41}$ & $18.76_{\pm0.58}$ & $14.56_{\pm0.50}$ & $12.68_{\pm0.64}$ & $11.14_{\pm0.42}$ & $11.09_{\pm0.41}$ \\
  \hline \\
\end{tabular}}
\caption{Test error on MNIST of a $10$-NN classifier in latent space, for a randomly initialised model with different latent dimensions.}
\label{table:knn_dim}
\end{table}

\section{Datasets}
\label{sec:datasets}
The MNIST dataset comprises handwritten samples of ten digits, split into $50,000$ training samples, $10,000$ validation samples, and $10,000$ test samples.
The Omniglot dataset comprises $20$ samples from each of $1623$ characters, grouped into $50$ different alphabets. While multiple splits are possible, we utilise a common method from previous work, with $15$ samples from each character in the training set, and $5$ in the test set.
In this case, we use the $50$ alphabets as the class labels for evaluation.
For all experimental runs, we train models for $10^5$ training steps.
In the sequential cases, this is equally divided between all classes, resulting in $10^4$ and $2\times10^3$ steps per class for MNIST and Omniglot, respectively.

\section{Experimental setup}
We used a single main setup for most of the experiments, with the exception of the external benchmarks, which are described separately in the following sections.
For all experiments, we train models for $10^5$ steps and report / plot the mean and standard deviation over $5$ different random seeds.
Each random seed for an experiment is trained using a single Tesla V100 GPU.
Each run takes approximately $1$ hour for the supervised SplitMNIST scenario, $3$ hours for the main unsupervised MNIST sequential run, and $10$ hours for Omniglot. The main bottleneck in run time is marginalising over all components, so this could be optimised in future implementations.

\subsection{Main setup}
\label{sec:main_setup}
\paragraph{MNIST}
For MNIST, we employed an MLP encoder with layer sizes of $\{1200, 600, 300, 150\}$ to form the shared representation; followed by a softmax layer for $\y$, with a maximum capacity of $K=25$ components; and a linear layer with $64$ output dimensions, for each component, to output the posterior parameters of the corresponding $32$-dimensional latent Gaussian.
Half of the output dimensions were used directly for the mean, while the other half were passed through a softplus activation to produce the variance.
The decoder consisted of two fully connected layers of $500$ dimensions, followed by a Bernoulli output layer for the reconstruction.
Both encoder and decoder networks employed ReLU nonlinearities, and training was performed using the ADAM optimiser with learning rate $10^{-3}$.

This architecture was obtained by performing a small hyperparameter sweep, also considering an encoder with two $500$-dimensional fully-connected layers, and a decoder with layer sizes $\{150, 300, 600, 1200\}$, but we observed only small differences in performance.

For dynamic expansion, we set the threshold for the log-likelihood (approximated by the ELBO) at $\cexp=-200$, and anything below this value is considered a poorly-explained sample and added to the buffer.
This was obtained by performing a sweep over values of $\{-100, 200, 400\}$, where this parameter can be use to directly modulate the expansion rate, and hence the capacity of the model (i.e. it is a way to automatically estimate the required number of components K). 
A fixed consolidation period of $100$ steps was used after each expansion before the model was re-eligible for expansion: this ensures that the model is able to learn from the data and fit a new component, and only flag poorly defined samples once learning has matured.
For all hyperparameter sweeps, model selection was performed using the validation set.

\paragraph{Omniglot}
For Omniglot, we employed an MLP encoder with layer sizes of $\{500, 500\}$ to form the shared representation; followed by a softmax layer for $\y$, with $K=25$ components; and a linear layer with $64$ output dimensions, for each component, to output the posterior parameters of the $32$-dimensional corresponding Gaussian.
Half of the output dimensions were used directly for the mean, while the other half were passed through a softplus activation to produce the variance.
The decoder consisted of one fully-connected layer of $500$ dimensions, followed by a Bernoulli output layer for the reconstruction.
Both encoder and decoder networks employed ReLU nonlinearities, and training was performed using the ADAM optimiser with learning rate $5\times10^{-4}$

For dynamic expansion, the same hyperparameters were used as in the MNIST case.

\subsection{Supervised continual learning benchmark}
\label{sec:supervised_setup}

For this external comparison, we employ the same hyperparameters as those reported in the previous work~\citep{hsu2018re, van2019three}.
The encoder comprised two fully-connected ReLU layers with $400$ dimensions, and we use a $100$-dimensional latent space with a capacity of $K=10$ components (matching the number of labels).
The decoder also comprises two fully-connected ReLU layer with $400$ dimensions, and Bernoulli outputs.
Dynamic expansion is performed in a supervised fashion (i.e. expanding when new class labels are introduced), and MGR is also used, with snapshotting at the end of each task.

\subsection{Unsupervised i.i.d learning benchmark}
\label{sec:clustering_setup}

For the external comparison, we employ the same hyperparameters as those reported in the previous work~\citep{joo2019dirichlet, nalisnick2017stick}.
For MNIST, the encoder comprised two fully-connected ReLU layers with $500$ dimensions, and a $50$-dimensional latent space with a maximum mixture capacity of $K=100$ components.
The decoder comprised a single fully-connected ReLU layer with $500$ dimensions, and Bernoulli outputs.
For Omniglot, the same architecture was used, but with a $100$-dimensional latent space, and with maximum $200$ mixture components.
By default, we employ dynamic expansion, and use MGR for the sequential learning case, using the ``dynamic'' snapshot approach.
Training was performed using the ADAM optimiser with learning rate $5\times10^{-4}$.

\end{document}